\documentclass[sigconf,authorversion]{acmart}
\usepackage{subcaption}

\AtBeginDocument{%
  }

\copyrightyear{2026}
\acmYear{2026}
\setcopyright{cc}
\setcctype{by}
\acmConference[SAC '26]{The 41st ACM/SIGAPP Symposium on Applied Computing}{March 23--27, 2026}{Thessaloniki, Greece}
\acmBooktitle{The 41st ACM/SIGAPP Symposium on Applied Computing (SAC '26), March 23--27, 2026, Thessaloniki, Greece}
\acmPrice{}
\acmDOI{10.1145/3748522.3779970}
\acmISBN{979-8-4007-2294-3/2026/03}

\begin{document}
\title{Heterogeneity in Multi-Robot Environmental Monitoring for Resolving Time-Conflicting Tasks}

\author{Connor York}
\orcid{0009-0001-8394-2671}
\affiliation{%
  \institution{University of Bristol}
  \city{Bristol}
  \country{United Kingdom}
}
\email{connor.york@bristol.ac.uk}

\author{Zachary R. Madin}
\orcid{0009-0005-0310-4504}
\affiliation{%
  \institution{University of Bristol}
  \city{Bristol}
  \country{United Kingdom}
}
\email{zachary.madin@bristol.ac.uk}

\author{Paul O'Dowd}
\orcid{0000-0002-1342-4246}
\affiliation{
  \institution{University of Bristol}
  \city{Bristol}
  \country{United Kingdom}
  }
\email{paul.odowd@bristol.ac.uk}

\author{Edmund R. Hunt}
\orcid{0000-0002-9647-124X}
\affiliation{
  \institution{University of Bristol}
  \city{Bristol}
  \country{United Kingdom}
  }
\email{edmund.hunt@bristol.ac.uk}

\renewcommand{\shortauthors}{York et al.}

\begin{abstract}
Multi-robot systems performing continuous tasks face a performance trade-off when interrupted by urgent, time-critical sub-tasks. We investigate this trade-off in a scenario where a team must balance area patrolling with locating an anomalous radio signal. To address this trade-off, we evaluate both behavioral heterogeneity through agent role specialization (``patrollers" and ``searchers") and sensing heterogeneity (i.e., only the searchers can sense the radio signal). Through simulation, we identify the Pareto-optimal trade-offs under varying team compositions, with behaviorally heterogeneous teams demonstrating the most balanced trade-offs in the majority of cases. When sensing capability is restricted, heterogeneous teams with half of the sensing-capable agents perform comparably to homogeneous teams, providing cost-saving rationale for restricting sensor payload deployment. Our findings demonstrate that pre-deployment role and sensing specialization are powerful design considerations for multi-robot systems facing time-conflicting tasks, where varying the degree of behavioral heterogeneity can tune system performance toward either task.

\end{abstract}

\ccsdesc[500]{Computer systems organization~Robotic autonomy}
\ccsdesc[500]{Computing methodologies~Multi-agent systems}
\ccsdesc[500]{Networks~Mobile and wireless security}

\keywords{Heterogeneity, Multi-Agent Systems, Multi-Robot Patrol, Collective Behavior, Anomaly Detection, Source Seeking}

\maketitle

\section{Introduction} 
Effective long-term environmental monitoring by autonomous systems is often punctuated by the need for rapid responses to transient, time-critical events. Multi-robot systems (MRS) are particularly well-suited to these dual demands, offering persistent and dynamic coverage that is not possible with static sensors (e.g., CCTV) or human teams. In the context of electromagnetic (EM) environment security, robots can be equipped with a suite of sensors to detect and measure EM signals from concealed devices. This allows them to not only locate any potential concealed devices (technical security), but also maintain long-term surveillance to prevent additional devices from being planted (physical security).

In this work, we formalize this challenge through a scenario that combines two competing tasks: a continuous \textbf{multi-robot patrol} (MRP) task, aimed at minimizing the idleness (revisit time) of key locations, and an urgent \textbf{source-seeking} task to locate an anomalous signal. Because a robot cannot perform both tasks simultaneously, a direct trade-off emerges: allocating more robots to search for the signal improves localization time but degrades patrol coverage, creating a potential security vulnerability.
We hypothesize that a system where only a subset of robots are pre-assigned to respond to the source-seeking task, whilst the rest remain allocated to the ongoing patrol task, will be more effective overall. We also hypothesize that sensing heterogeneity, whereby a subset of robots are equipped with the sensor(s) necessary to detect and engage with the source-seeking task, might motivate a more cost-effective multi-robot system. We investigate these hypotheses through a parameter sweep of agent role and sensing distributions.

We draw inspiration from heterogeneity in nature, where individual specialization (both morphological and behavioral) has been shown to confer fitness advantages \cite{o2021functional,dall2012evolutionary, svanback2003morphology}. Heterogeneity has been investigated in previous robotic systems, with behavioral heterogeneity being shown to increase performance in foraging tasks \cite{york2024shaping, debie2025task} and decision making \cite{zakir2024heterogeneity}.

\begin{figure}[ht]
    \centering
    \begin{minipage}[b]{0.49\linewidth}
        \centering
        \includegraphics[width=\linewidth]{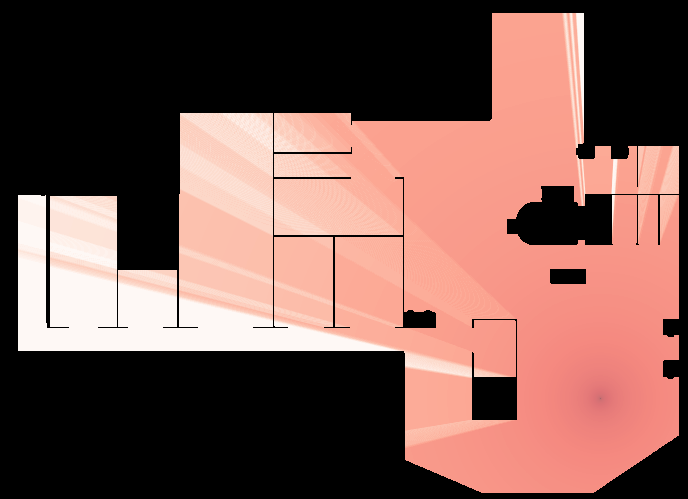}
    \end{minipage}
    \hfill
    \begin{minipage}[b]{0.49\linewidth}
        \centering
        \includegraphics[width=\linewidth]{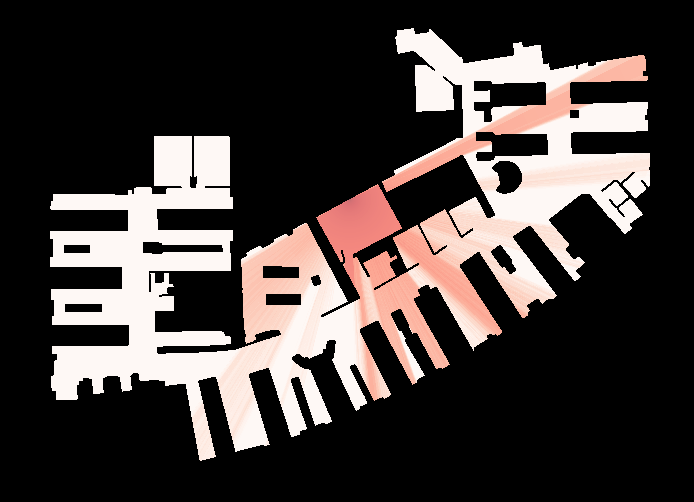}
    \end{minipage}
    \caption{The two simulation maps: Cumberland (left) and Office (right), each showing the propagation of one of their three tested signal source locations in red gradient.}
    \label{fig:maps}
\end{figure}

\section{Methodology}
We implement a simulation of mobile robots assigned two concurrent tasks with a conflict in the time domain.  First, the robots must patrol an indoor environment and minimize the time that predetermined nodes on the route are unvisited, evaluated as \emph{idleness}. Second, an anomaly occurs in the environment (e.g., an espionage device emitting a signal) which must be spatially localized, evaluated as the \emph{time to find (TTF)}.  To balance this trade-off we split the agents into two classes, ``\emph{patroller}" agents which are only able to undertake the patrolling task, and ``\emph{searcher}" agents whose baseline behavior is also to patrol, but will search for the anomalous signal source when it is detected until it has been located. The proportion of ``\emph{searcher}" robots constitutes the degree of heterogeneity of the robotic system, and is varied as the subject of investigation.

\subsection{Simulator} 
For this scenario, we model the environment as a two dimensional grid-world, with cells either being free space or entirely occupied (e.g. by a wall). We employ two maps shown in Fig. \ref{fig:maps}: ``Cumberland" taken from \textit{patrolling\_sim} \cite{portugal2019ros}, and ``Office" created from LIDAR scans of a university office. Robot speeds were bound to $v_{max}=0.4 m/s$, and the simulator configured such that 1 timestep can be considered 1 second of real time.

Simulated agents are provided with absolute position information and a map of the environment to navigate by. Every timestep agents communicate to any other agent within a specified range: their current position, believed global best estimate of the target emitter's position, and whether the anomaly has been found. The communication range is varied as an experiment parameter (see Section \ref{subsec:config}). The patrol route is defined as a graph by pre-set inter-connected nodes throughout the environment. Agents begin each trial evenly distributed across the nodes.  The chosen proportion of ``\emph{searcher}" agents are randomly selected from within all agents with equal probability. The anomalous signal appears at a random time between 400 and 600 seconds into the trial, and remains active until it has been ``found" by an agent (see Section \ref{subsec:Signal}). The anomaly's appearance is delayed to give a comparable starting condition across all trials by allowing patrolling to reach a stable state. This minimizes the effect of agents initializing closer to the signal source, and allows idleness to stabilize to a baseline.

\subsection{Signal Propagation}\label{subsec:Signal}

In order for the robot to spatially localize an EM signal emitting device, we model the wireless signal with a perceivable power decay over distance, using the multi-wall loss model from \cite{capulli2006path}, generating a ``signal map" as shown in Fig. \ref{fig:maps} shaded in red.
It begins with the free space path loss formula (\ref{eq:fpl}) and adds a multi-wall loss component $M_w$ (\ref{eq:Mw}). For our experiments we define the source as a transmitter with a power $P_{dBm}$ of 20 dBm (100 mW) and frequency $f$ of 2.4 GHz, which are common transmission parameters for wireless communications:

\begin{equation}
    l_{dB} = 20log_{10}(d) + 20log_{10}(f) + 20log_{10}\left(\frac{4\pi}{c}\right) + M_w
    \label{eq:fpl}
\end{equation}

\begin{equation}
    M_w = \sum\limits_{j=1}^{J}k_{wj}l_j
    \label{eq:Mw}
\end{equation}

\begin{equation}
    RSSI = P_{dB} - l_{dB}
    \label{eq:RSSI}
\end{equation}

The multi-wall component consists of $k_{wj}$ being the number of walls crossed of type $j$ and the attenuation $l_j$ due to walls of type $j$. We use the value $l_j$ = 4 for all walls \cite{obeidat2018indoor}.

Equation (\ref{eq:fpl}) was calculated for each open cell, with each occupied cell in a straight line from there to the source position counting as a single wall. As the measured signal is represented by RSSI (received signal strength indicator) we are able to define a value that is sufficient to mark the source as ``found" when measured by an agent, without having knowledge of the underlying signal decay. For experiments this value was set to $-20$ dBm, equivalent to 1 m free space path loss in (\ref{eq:fpl}). We generated signal propagation maps for three different transmitter locations per map, with locations chosen to provide different propagation effects (e.g. open space, corner of map, heavily obfuscated) representing different challenges to the robotic system. Figure \ref{fig:maps} shows each map overlaid with one of their three generated signal maps.  

\subsection{Multi-Robot Patrol}\label{subsec:meth_mrp}
To achieve multi-robot patrolling (MRP), we adopt State Exchange Bayesian Strategy (SEBS) \cite{portugal2013distributed}. Upon arrival to a node, robots select their next goal from the neighboring nodes based on perceived node idleness and the communicated intentions of other robots. 

\subsection{Agent Search}
Agent search begins when a searcher agent either detects the signal itself, or receives communicated signal information from another agent. The following algorithms provide a potential goal for the agents to navigate to via A*. If the proposed goal intersects with an obstacle, it is moved to the first free space on the other side of the obstacle, continuing in the same direction of travel. If the goal is the result of agent repulsion (in PSO or HC-PSO) the robot instead stays still for that update.

\begin{figure*}[ht]
    \centering
    \includegraphics[width=0.9\textwidth]{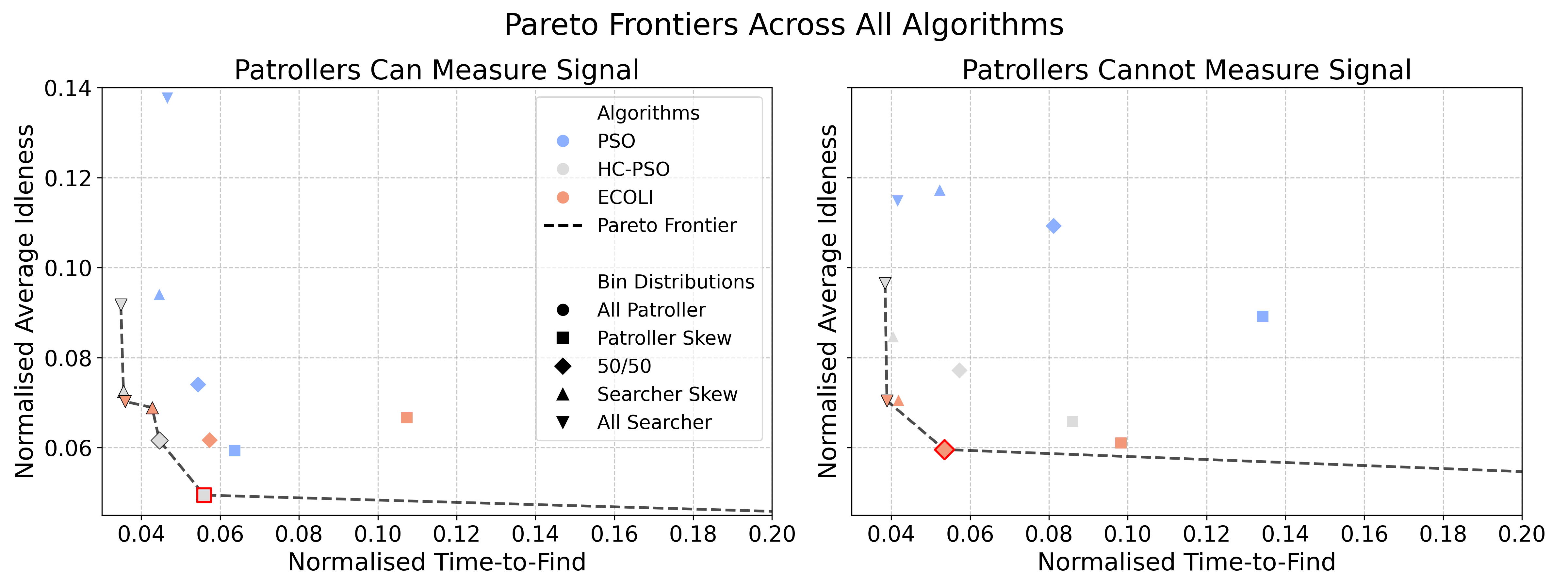}
    \caption{Pareto Frontier across all algorithms/distribution medians, ``knee point" bounded in red, and frontier points are linked by dashed line. Axis is cropped to exclude ``\emph{All Patroller}" groups clustered near (1.0, 0.0).}
    \label{fig:pareto}
\end{figure*}

\subsubsection{PSO}
Our implementation of PSO is based on Charged PSO \cite{blackwell2002dynamic} where the standard PSO velocity update has an added repulsion between agents with the intent to maintain continuous agent mobility until a suitably strong signal is measured. Agents are repulsed from the other robot's communicated positions, with magnitude decreasing linearly over distance. The cognitive and social coefficients are set as $c_1 = 1.0, \space c_2=2.5$. 

\subsubsection{E. Coli Chemotaxis (ECOLI)}
We implement an \textit{E. coli} chemotaxis algorithm inspired by \cite{holland1996some}. This is a reactive approach where the agent compares its current signal reading to the one prior. If the current reading is worse than the prior, it reverses course by $180^\circ$, otherwise continuing on. To maintain diversity in the search, after the initial course adjustment there is a 50\% chance of adjusting the heading by $\pm45^\circ$ (25\% chance for either direction). 

\subsubsection{Hybrid-Chemotactic PSO (HC-PSO)}
The third algorithm (HC-PSO) combines the previous two with the majority of the agents searching via PSO, whilst the agent(s) responsible for their global-best (highest strength) signal reading searches via ECOLI. 

\subsection{Experiment Configurations}\label{subsec:config}
Simulations were run for 2000 timesteps on two maps (Fig. \ref{fig:maps}) with three source locations each. Each parameter configuration (see Table \ref{tab:parameters}) was repeated 15 times. We measured node idleness (averaged from timestep 250 onward, allowing patrolling to stabilize) and the time taken to find the source (TTF).

\begin{table}[ht]
    \centering
    \caption{Experimental parameters}
    \label{tab:parameters}
    \setlength\tabcolsep{3pt}
    \begin{tabular}{l l}
    \toprule
     \textbf{Parameter} & \textbf{Values} \\
    \midrule
    Group Size ($N$) & Cumberland: \{6, 12\}, Office: \{12, 20\} \\
    Number of Searchers ($k$) & \{0, 1, 2, \ldots, $N$\} \\
    Search Algorithm & PSO, HC-PSO, ECOLI \\
    Communication Range & 1.5 m, 2.5 m, 4 m, 6 m, 8 m, Global \\
    Patrollers can -\\
    - Measure Signal (\textit{PM}) & True, False\\
    \bottomrule
    \end{tabular}
\end{table}

\section{Results \& Discussion}
Results were grouped by map and $N$ agents, with idleness and TTF performance min-max normalized within those groups such that 0.0 represents the best performance. The normalized results were then grouped into relative \textbf{patroller:searcher} distributions shown in Table \ref{table:ratios}. This normalization and grouping allows us to investigate the effects of relative distributions regardless of agent count and environment. Pareto-front analysis was used to investigate the \textbf{idleness:TTF trade-off} offered by each agent distribution. The optimal balance between idleness and TTF minimization is represented by the ``knee point", or the point on the Pareto front closest to the origin. To verify our results we used bootstrap validation: 10,000 samples were drawn with replacement from the full dataset, recalculating medians and Pareto points to verify the rate at which each ``knee point" occurs and each point is Pareto optimal. Statistical comparisons between the idleness/ TTF of configurations were performed using Mann-Whitney U tests with rank-biserial correlation as the effect size measure.

Figure \ref{fig:pareto} shows the results of the Pareto-front analysis on the distribution medians across all algorithms, split by \textit{PM} True/False. 
When \textit{PM} = True: the ``knee point" configuration is HC-PSO with a \emph{Patroller Skew} distribution in 52\% of bootstrapped samples (with this configuration being Pareto optimal in 99\% of samples), followed by the same algorithm with a \emph{50/50} distribution in 28\% of samples. The knee points for individual algorithms are: \emph{Patroller Skew} for PSO (75\% of samples), \emph{Patroller Skew} for HC-PSO (65\% of samples), and homogeneous \emph{All Searcher} for ECOLI (55\% of samples). By including a minority of searchers, \emph{Searcher Skew} for HC-PSO sees a 22\% decrease in median TTF ($U=72101, p<0.001, r=0.152$) with no significant increase in idleness ($U=87143, p=0.300, r=-0.025$). This shows that when specialization does not restrict the amount of information available to the system (i.e., \textit{PM} = True), behaviorally heterogeneous teams can balance both objectives without large performance sacrifices, enabling tuning of performance along the Pareto front by simple, high-level changes to team role composition. 

Alternatively, when \textit{PM} = False: the balanced knee point is ECOLI with one of two distributions: \emph{All Searcher} is the knee point in 39\% of bootstrapped samples and Pareto optimal in 81\% of samples, whilst \emph{50/50} is the knee point in 37\% of samples but Pareto optimal in 97\% of samples. The knee points for individual algorithms are: \emph{All Searcher} for PSO (69\% of samples), \emph{Searcher Skew} for HC-PSO (55\% of samples), and again \emph{All Searcher} and \emph{50/50} for ECOLI (39\% and 38\% of samples respectively). 
Whilst in this case heterogeneity often presents more of a trade-off due to also decreasing sensing capability across the system, the ECOLI algorithm is able to maintain similar task balance with a \emph{50/50} split compared to the homogeneous \emph{All Searcher} case. This demonstrates system costs can be effectively reduced through a heterogeneous team with limited sensor deployment, whilst maintaining effective balance between the two tasks.

\begin{table}[]
    \centering
    \caption{\textbf{Patroller:Searcher} distribution groupings by $k$ searcher agents}
    \label{table:ratios}
    \setlength\tabcolsep{3pt}
    \begin{tabular}{l ccccc}
    \toprule
     \textbf{\textit{N}}  & \textbf{All Patrol} & \textbf{Patrol Skew} & \textbf{50/50} & \textbf{Search Skew} & \textbf{All Search}\\
    \midrule
    6  & 0              & 1, 2            & 3     & 4, 5           & 6             \\
    12 & 0              & $[1,5]$        & 6     & $[7,11]$      & 12            \\
    20 & 0              & $[1,9]$        & 10    & $[11,19]$     & 20           \\
    \bottomrule
    \end{tabular}
\end{table}

\begin{figure}[h]
    \centering
    \includegraphics[width=0.8\linewidth]{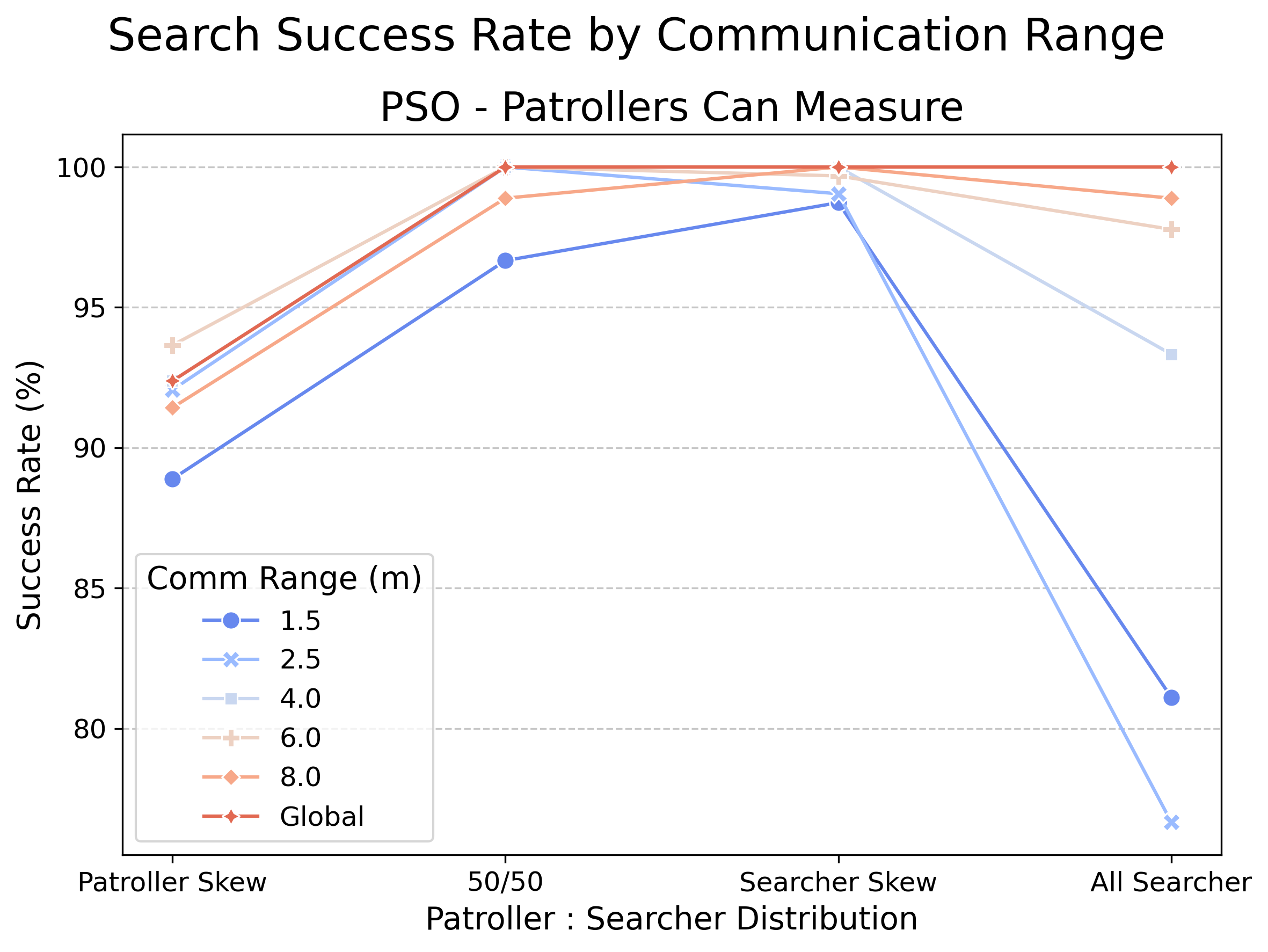}
    \caption{PSO search success rate for each agent distribution as communication range decreases. Data from 6 agents on ``Cumberland" map and 12 agents on ``Office" map, and \textit{PM} = True.}
    \label{fig:commrange}
\end{figure}

Figure \ref{fig:commrange} shows the decrease in search success rates when communication is restricted for PSO (the only algorithm requiring communication). We see a significant increase in search success rate through the inclusion of a minority of patroller agents for both cases (\textit{PM} = True or False), most notably at the lower communication ranges. For example, at 2.5 m communication range when \textit{PM} = True, \emph{Searcher Skew} has a 99\% search success rate versus 77\% for \emph{All Searcher} (Fisher's exact test: $OR=31.7$, $p<0.001$). This is because as the patroller agents maintain regular movement throughout the environment without converging on signal maxima, they are able to transfer information to any isolated robots and enable them to aid with the search. 
Whilst this phenomenon only appears for cases where communication is both necessary and restricted, it shows an emergent advantage of heterogeneity. This highlights how simple adjustments to team heterogeneity can yield unexpected benefits for system performance, providing further credence to the consideration of heterogeneity when deploying multi-robot systems.

\section{Conclusion}
When tasked with long-term environmental monitoring, multi-robot systems must manage their performance when interrupted by a short-term sub-task, as to focus on one requires sacrificing performance in the other. We contextualize this scenario as a security task requiring continuous patrolling of the environment to monitor for physical intrusion, whilst reacting to and locating any anomalous signals that may appear. We have demonstrated that behavioral heterogeneity through agent role specialization is an effective way to balance performance between time-conflicting tasks, whilst also providing a simple, high-level approach to tune the system to prioritize either goal through controlling pre-assigned role distributions. We extended our findings to sensing heterogeneity, showing that optimal balanced performance can be maintained even when only half of the swarm is equipped with anomaly sensing capabilities, providing a cost-saving rationale for restricting sensor deployment when designing multi-robot systems. Finally, we note an emergent advantage of behavioral heterogeneity in supporting inter-agent information transfer when communication is constrained, increasing search success rate.

These results demonstrate that agent heterogeneity through pre-deployment role or sensing specialization should not be overlooked when designing multi-robot systems tasked with time-conflicting objectives, and provide a baseline for further improvement through more complex strategies.
Future work will focus on validation of our findings through robotic trials using a real signal for the agents to follow, and investigating whether strategies where heterogeneous behaviors adapt to environmental information can outperform static deployments.

\begin{acks}
CY is supported by UK FCDO Services. ZRM is supported by a UoB PhD Scholarship. ERH is supported by the Royal Academy of Engineering under the Research Fellowship program.
\end{acks}

\bibliographystyle{ACM-Reference-Format}
\bibliography{references}

\end{document}